\documentclass[conference]{IEEEtran}
\IEEEoverridecommandlockouts
\usepackage{cite}
\usepackage{amsmath,amssymb,amsfonts}
\usepackage{algorithmic}
\usepackage{graphicx}
\usepackage{booktabs}
\usepackage{hyperref}
\usepackage{textcomp}
\usepackage{xcolor}
\usepackage{subcaption}
\usepackage{xcolor} % for color
\def\BibTeX{{\rm B\kern-.05em{\sc i\kern-.025em b}\kern-.08em
    T\kern-.1667em\lower.7ex\hbox{E}\kern-.125emX}}
\begin{document}

% Create a separate title page for the copyright statement
\begin{titlepage}
\thispagestyle{empty} % No headers or footers on this page
\vspace*{\fill} % Vertically center the content on the page
\begin{center}
    \textbf{Copyright Statement}\\[1cm]
    This paper has been accepted for publication in the IEEE MLISE 2024 conference \\

    © 2024 IEEE. Personal use of this material is permitted. Permission from IEEE must be obtained for all other uses, in any current or future media, including reprinting/republishing this material for advertising or promotional purposes, creating new collective works, for resale or redistribution to servers or lists, or reuse of any copyrighted component of this work in other works.
\end{center}
\vspace*{\fill}
\end{titlepage}

\title{CU-Net: a U-Net architecture for efficient brain-tumor segmentation on BraTS 2019 dataset}

\author{
    \IEEEauthorblockN{Qimin Zhang\textsuperscript{1}\textsuperscript{*}}
    \IEEEauthorblockA{
        \textit{Department of Biostatistics} \\
        \textit{Columbia University}\\
        New York, NY 10032, USA \\
        qimin.zhang@columbia.edu
    }
    \and
    \IEEEauthorblockN{Weiwei Qi\textsuperscript{1}\textsuperscript{*}}
    \IEEEauthorblockA{
        \textit{Department of Biostatistics} \\
        \textit{Columbia University}\\
        New York, NY 10032, USA \\
        wq2151@columbia.edu
    }
    \and
    \IEEEauthorblockN{Huili Zheng\textsuperscript{2}}
    \IEEEauthorblockA{
        \textit{Department of Biostatistics} \\
        \textit{Columbia University}\\
        New York, NY 10032, USA \\
        hz2710@caa.columbia.edu
    }
    \and
    \IEEEauthorblockN{Xinyu Shen\textsuperscript{2}}
    \IEEEauthorblockA{
        \textit{Department of Biostatistics} \\
        \textit{Columbia University}\\
        New York, NY 10032, USA \\
        xs2384@columbia.edu
    }
}

\maketitle

\begin{abstract}
Accurately segmenting brain tumors from MRI scans is important for developing effective treatment plans and improving patient outcomes. This study introduces a new implementation of the Columbia-University-Net (CU-Net) architecture for brain tumor segmentation using the BraTS 2019 dataset. The CU-Net model has a symmetrical U-shaped structure and uses convolutional layers, max pooling, and upsampling operations to achieve high-resolution segmentation. Our CU-Net model achieved a Dice score of 82.41\%, surpassing two other state-of-the-art models. This improvement in segmentation accuracy highlights the robustness and effectiveness of the model, which helps to accurately delineate tumor boundaries, which is crucial for surgical planning and radiation therapy, and ultimately has the potential to improve patient outcomes.
\end{abstract}


\section{Introduction}

Brain tumors\cite{herholz2012brain} are clumps of abnormal cells that grow and multiply uncontrollably within the brain. If not detected and treated promptly, these tumors can have devastating effects on the body. Symptoms of brain tumors vary greatly depending on their location, size, and type (benign or malignant), and can include headaches, seizures, vision problems, cognitive impairment, and even paralysis. Malignant brain tumors are particularly dangerous because they invade and destroy surrounding healthy brain tissue and can spread to other parts of the body\cite{sampson2017immunotherapy}.

Magnetic resonance imaging (MRI)\cite{Shen2024Harnessing} is an important diagnostic method for detecting and evaluating brain tumors\cite{herholz2012brain}. Utilizing powerful magnetic fields and radio waves, MRI scans produce highly detailed images of the brain's anatomy, helping to accurately identify the presence, location, and extent of a tumor. These images are essential for treatment planning, providing important insights into the tumor's size, shape, and its relationship to vital brain structures. Additionally, MRI scans play a vital role in monitoring tumor progression or regression during and after therapeutic interventions\cite{wen2010updated}.

Early and accurate detection of brain tumors is critical for effective treatment and improved patient outcomes. However, manually segmenting brain tumors from MRI images is a challenging and time-consuming task that requires significant expertise and is susceptible to inter-observer variability. The process involves carefully delineating the boundaries of the tumor from the surrounding healthy brain tissue, which is particularly difficult for irregularly shaped or diffuse tumors. Therefore, the development of robust automated models for brain tumor segmentation is critical for efficient and reliable diagnosis and treatment planning. Automated segmentation models\cite{chen2023learning} can consistently and accurately identify and delineate tumors, greatly reducing the time and effort required for manual analysis and minimizing the potential for human error or discrepancies. Additionally, these models have the potential to detect subtle or complex tumor patterns that human experts may miss, enabling earlier and more accurate diagnoses.

Machine learning (ML) and deep learning (DL) techniques have revolutionized various fields, including regression\cite{zhou2024predict, yu2024credit}, image recognition\cite{xin2024parameter, zhou2024research, li2022quantized}, natural language processing\cite{mo2024large, luo2024enhancing}, and robotics\cite{liu2024enhanced}, with key applications in medicine for classification\cite{read2023prediction} and segmentation tasks. Convolutional neural networks (CNNs)\cite{lecun1989backpropagation} are particularly effective in automatically learning hierarchical features from raw image data, allowing for accurate and efficient tumor segmentation. This data-driven approach excels in medical image classification and segmentation\cite{chen2023generative} by capturing intricate details and subtle variations that traditional handcrafted features may miss. Deep learning models have consistently outperformed traditional methods in managing the complexity and variability of brain tumor images. Enhancements such as attention mechanisms and multimodal data fusion further improve the accuracy and robustness of these models by focusing on relevant image regions and integrating data from various imaging modalities (e.g., MRI, CT scans). In addition, recurrent neural networks (RNNs)\cite{jiang2021recurrent} are also used in healthcare for tasks such as patient monitoring, disease prediction, and medical data analysis, highlighting the broad applicability and impact of deep learning in medicine.

\section{Data}
There are a large number of medical image databases available that provide a valuable resource for medical imaging research and development. These databases include various imaging modalities such as MRI, CT scans, and X-rays, which are essential for tasks such as image segmentation, classification, and diagnostic analysis\cite{chen2024bimcv}.

\subsubsection{Dataset}
\begin{figure}[hbt!]
    \centering
    \includegraphics[width=0.9\linewidth]{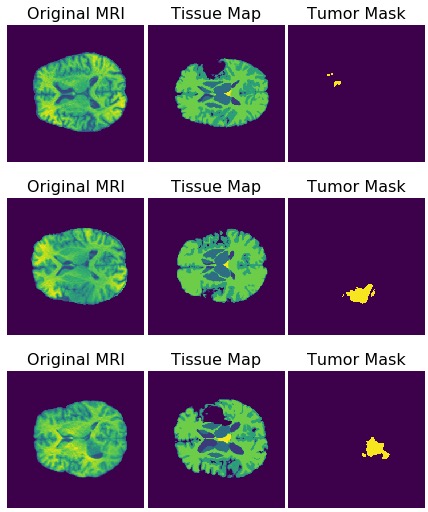}
    \caption{Sample BraTS 2019 images}
    \label{fig:brats2019_samples}
\end{figure}

We used the BraTS dataset of MRI\cite{menze2014multimodal}, which is by far the largest and most comprehensive dataset used for brain-tumor segmentation. All BraTS multimodal scans are available as NIfTI files and include native T1-weighted, post-contrast T1-weighted, T2-weighted, and T2 Fluid Attenuated Inversion Recovery (FLAIR) volumes. All the images are segmented manually by experienced neuro-radiologists. Specifically, we used the BraTS 2019 dataset to train our network. Sample BraTS 2019 images are shown in Fig.~\ref{fig:brats2019_samples}.

\subsubsection{Dataset Split and Pre-Processing}
The BraTS 2019 dataset provides 335 subjects, of which 80\% was used as a training set, 10\% as the validation set, and the remaining 10\% as the test. As the tumor segmentation results are provided in the binary form, the tumor mask needed to be pre-processed to a binary mask before training the segmentation network. 

\section{Methods}
\subsection{Convolutional Neural Network}
Convolutional Neural Networks (CNNs) are a class of deep learning models that have revolutionized image recognition and classification tasks due to their ability to automatically learn hierarchical representations from raw input data. At the core of a CNN are convolutional layers, which apply a set of learnable filters (or kernels) to the input image to produce feature maps. The mathematical operation performed by a convolutional layer is defined as:

\begin{equation}
(f * g)(t) = \int_{-\infty}^{\infty} f(\tau) g(t - \tau) d\tau,
\end{equation}

where \( f \) is the input image and \( g \) is the filter. This operation captures local dependencies by detecting edges, textures, and other patterns. Following the convolutional layers are pooling layers, which reduce the spatial dimensions of the feature maps, enhancing computational efficiency and providing a form of translation invariance. The pooling operation is often a max-pooling function, defined as \( y = \max(x_i) \), where \( x_i \) are the elements within a pooling window. These operations, combined with non-linear activation functions like ReLU \( f(x) = \max(0, x) \), enable CNNs to capture complex patterns and structures in images\cite{lecun1998gradient}.

\subsection{CU-Net: Segmentation Model}

\begin{figure}[hbtp]
    \centering
    \hspace*{-1.1cm} % Adjust this value to move the image to the left
    \includegraphics[width=1.26\linewidth]{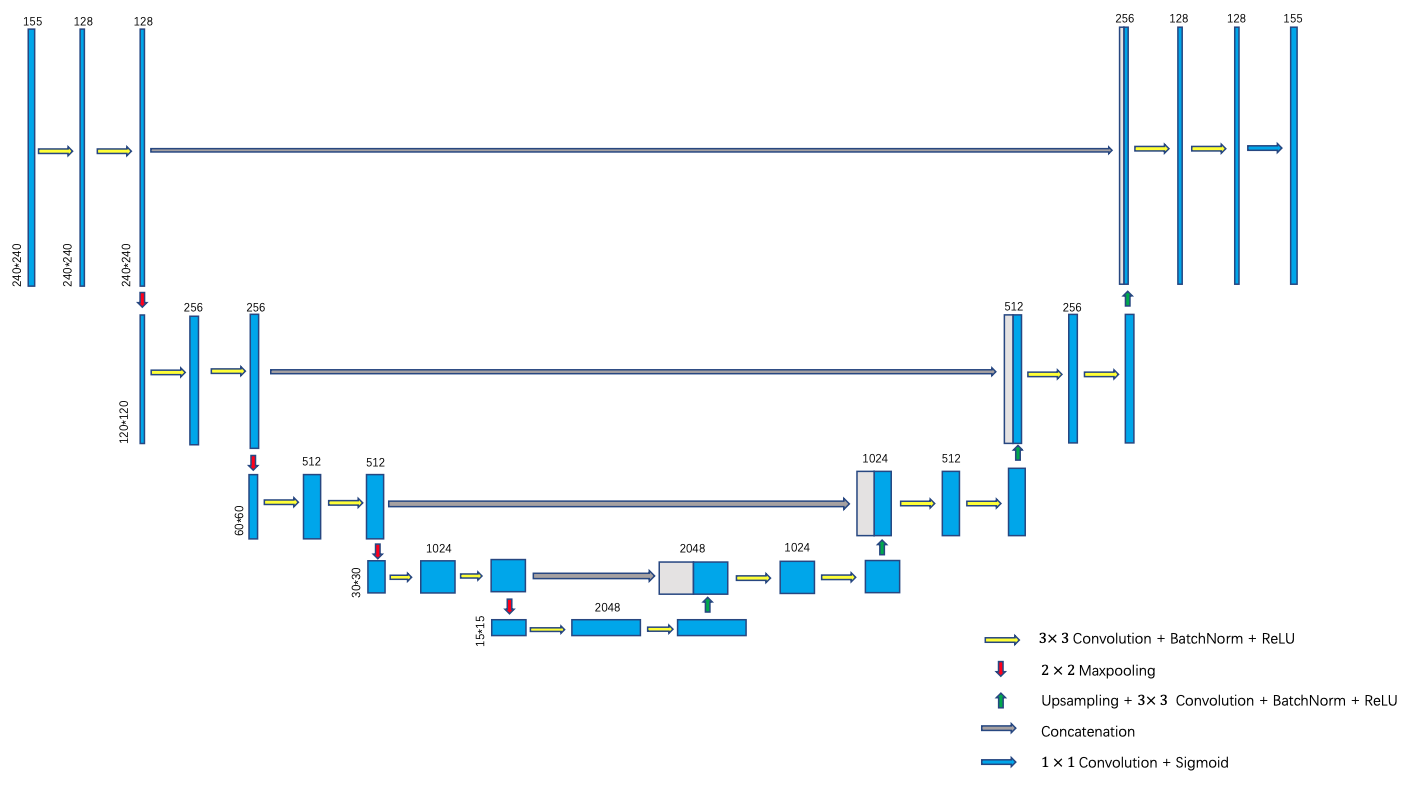}
    \caption{CU-Net Architecture}
    \label{fig:unet}
\end{figure}

We implemented Columbia-University-Net (CU-Net) as the segmentation model, which is a convolutional neural network (CNN) with adding a normal contracting network by
successive layers, where pooling operators are replaced by upsampling operators. The CU-Net architecture has a large number of feature channels in the upsampling part, which helps the network to extract more information and lead to higher resolution outputs. In addition, in order to localize, high resolution features from the contracting path are combined with the upsampled output\cite{ronneberger2015}. Therefore, the output images are with high resolution. The detailed architecture of our model is shown in Fig.~\ref{fig:unet}, which features a symmetric U-shaped structure with a contracting path, a bottleneck, and an expansive path. 

\subsubsection{Contracting Path}
Begins with an input layer for images of size $240 \times 240 \times 155$. It uses blocks of $3 \times 3$ convolutions followed by batch normalization and ReLU activation. Each convolutional block is followed by a $2 \times 2$ max pooling layer, doubling the feature map depth from 128 to 1024 while reducing spatial dimensions.

\subsubsection{Bottleneck}
The lowest layer, without pooling, prepares features for the expansive path.

\subsubsection{Expansive Path}
Features spatial up-sampling and $3 \times 3$ convolutions, with batch normalization and ReLU. Up-sampled maps are concatenated with feature maps from the contracting path, halving feature depth progressively. 

\subsubsection{Output Layer}
Concludes with a $1 \times 1$ convolution and a sigmoid activation to produce the segmented image.

\subsection{Loss Function}
For the segmentation model, we used Binary Cross-Entropy (BCE) as the loss function as follows:

\begin{equation}
BCE = -(ylog(p)+(1-y)log(1-p))
\end{equation}
where $y$ is the ground truth label and $p$ is the predicted label.

\subsection{Evaluation Metric}
Dice Score was used to measure the similarity between two sets:

\begin{equation}
Dice\;Score = \frac{2|X \cap Y|}{|X + Y|}
\end{equation}
where $X$ and $Y$ are two sets. We here used it to evaluate model performance, where 0 represents the worst performance and 1 represents the best. The final dice scores were evaluated on test set, i.e., 10\% of the original data.

\section{Experiments and Results}

\subsection{Experiments}

We implemented our model in PyTorch and trained it on Google Cloud Platform with an NVIDIA Tesla K80 GPU. The segmentation network was trained on 80\% of the BraTS19 data (269 subjects), and its performance was continuously evaluated based on the Dice score computed on the validation set (10\% of the data). The epoch that yielded the highest Dice score on the validation set was identified, and the model configuration from this epoch was selected as the optimal model. This best-performing model was subsequently applied to the testing set (10\% of the data) to assess its generalizability and effectiveness.

\subsection{Results}

\begin{figure}[htbp]
\centering
% First row of images
\begin{subfigure}[b]{0.8\linewidth}
    \includegraphics[width=\linewidth]{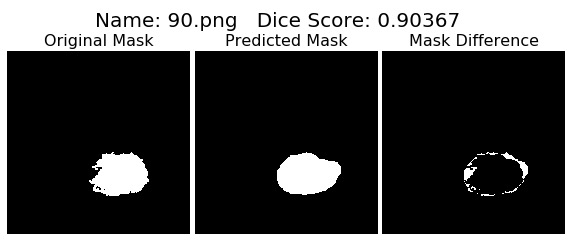}
    \label{fig:sub1}
\end{subfigure}

\begin{subfigure}[b]{0.8\linewidth}
    \includegraphics[width=\linewidth]{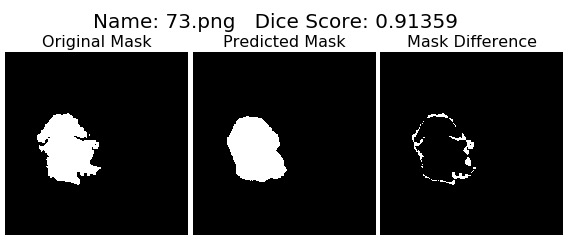}
    \label{fig:sub2}
\end{subfigure}

\begin{subfigure}[b]{0.8\linewidth}
    \includegraphics[width=\linewidth]{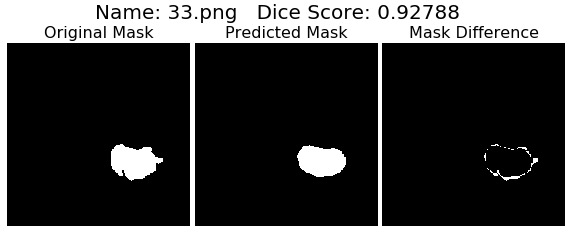}
    \label{fig:sub4}
\end{subfigure}

\caption{Comparison of Mask Differences}
\label{fig:mask_diff}
\end{figure}

We compared dice Score evaluation results of our model to other two state-of-art models in Tab~\ref{tab:dice_scores}. Also, samples are provided to visualize the performance of our model in Fig.~\ref{fig:mask_diff}. 

\begin{table}[htbp]
\caption{Dice Scores of Different U-Nets}
\begin{center}
\begin{tabular}{|c|c|}
\hline
\textbf{Model} & \textbf{Dice Score} \\
\hline
Swin UNet\cite{cao2022swin} & 81.45\% \\
\hline
TransUNet\cite{chen2021transunet} & 82.31\% \\
\hline
\textbf{Our CUNet} & \textbf{82.41\%} \\
\hline
\end{tabular}
\label{tab:dice_scores}
\end{center}
\end{table}

\section{Conclusion}

In this paper, we trained an efficient U-Net called CU-Net for brain-tumor segmentation with dice score 82.41\% on BraTS 2019 data, which surpasses the scores reported in two recent studies. Specifically, Swin UNet\cite{cao2022swin} reported a Dice score of 81.45\%, and TransUNet\cite{chen2021transunet} achieved 82.31\%, as documented in their respective publications.

\section{Discussion}

The model introduced in this study represents a major advance in brain tumor segmentation. The high Dice scores obtained by our model indicate its enhanced accuracy and effectiveness in delineating brain tumors, an important aspect of diagnostic imaging in neuro-oncology\cite{zhang2020enhanced}. Improved segmentation accuracy is critical not only for confirming the presence of a tumor but also for defining its precise anatomical boundaries, which is critical for subsequent medical procedures\cite{li2019advances}.

This advancement has significant implications for clinical practice, particularly in planning and executing neurosurgical interventions and personalized treatment options. Accurate tumor delineation is critical to developing a surgical plan designed to maximize tumor removal while minimizing risks to critical brain functions\cite{kumar2021impact}. Furthermore, precise segmentation directly affects the efficacy of radiation therapy, where the precise contours of the tumor determine the dose and accuracy of the radiation beam\cite{chen2018radiation}. Therefore, our model significantly contributes to improving the quality of neuro-oncology patient care, hopefully improving treatment outcomes and potentially improving patient survival and quality of life after treatment\cite{singh2022role} if supported by cutting-edge technologies\cite{lin2020touch}. In addition, due to the challenges caused by insufficient labeled data in medical imaging, self-supervised learning\cite{zhao2024optimization} can leverage large amounts of unlabeled data to significantly enhance medical imaging tasks such as tumor segmentation, disease detection, and anomaly detection, further improving the accuracy and robustness of our model.

\bibliography{report}
\bibliographystyle{IEEEtran}

\end{document}